\documentclass[11pt,a4paper]{article}
\usepackage[hyperref]{acl2020}
\usepackage{nert}
\usepackage{times}
\usepackage{latexsym}
\usepackage{float}
\usepackage{svg}
\usepackage{tipa}
\usepackage{layouts}
\usepackage{graphicx}

\usepackage{microtype}

\aclfinalcopy 

\title{Bhāṣācitra: Visualising the dialect geography of South Asia}

\author{Aryaman Arora \\
  Georgetown University \\
  \texttt{aa2190@georgetown.edu}\\
  \And
  Adam Farris \\
  San Mateo High School \\
  \texttt{adamfarris@gmail.com} \\\AND
  Gopalakrishnan R \\
  EFL University, Hyderabad \\
  \texttt{gopalkrishnan11251@gmail.com}\\\And
  Samopriya Basu \\
  University of North Carolina -- Chapel Hill \\
  \texttt{sampr0b@live.unc.edu} \\
}

\date{}

\begin{document}
\maketitle
\begin{abstract}
We present Bhāṣācitra,\footnote{From Sanskrit \textit{bhāṣā} `language' + \textit{citra} `ornament, appearence'; lit.~`language map'.} a dialect mapping system for South Asia built on a database of linguistic studies of languages of the region annotated for topic and location data. We analyse language coverage and look towards applications to typology by visualising example datasets. The application is not only meant to be useful for feature mapping, but also serves as a new kind of interactive bibliography for linguists of South Asian languages.
\end{abstract}

\section{Introduction}
South Asia is extremely linguistically diverse. There is a common saying illustrating this diversity, present in several languages of the region; it is given in Hindi below.
\begin{quote}
    \textit{kos kos par pānī badle, cār kos par bānī.}\\
    `The taste of water changes every mile, and the language every four.'
\end{quote}
One issue with this vast scale of diversity is the difficulty it poses for linguists in collecting and cataloguing linguistic data, which further impedes comprehensive typological analysis. India alone contains known living speakers of 461 languages \citep{ethnologue}.\footnote{But note \citet{asher2008language}: ``It is impossible to be at all precise about either the number of languages spoken in the region or the number of speakers of each.''} It is also difficult to assess the availability of linguistic literature for all of these languages, leading to gaps in the typological databases we end up compiling; print linguistic bibliographies for the region become outdated as new work is published and do not encode useful metadata, such as the specific dialect studied in each work or the linguistic features studied.

In this paper we present \textbf{Bhāṣācitra}, a database of linguistic sources for South Asian languages that we have compiled and annotated, as well as a dialect mapping and visualising system built from the location data extracted from those sources. Currently it includes 1104 labelled sources covering 311 lects. The site is online at \url{http://aryamanarora.github.io/bhasacitra}.

\section{Background and related work}

Dialects\footnote{\textit{Dialect} for the purposes of this paper refers to any speech variety. South Asia as a region is prone, due to geographical and historical factors, to fuzzy boundaries between speech varieties. The situation is best explained by \citet{deo2018dialects} in describing the distribution of Indo-Aryan as ``sociolinguistically rich and complex, characterized by plurilinguality and dialect continua spread over large regions spanning multiple languages''.} are defined by \textit{isoglosses}, geographical boundaries separating linguistic features. The mapping of dialect geography is a well-established problem in linguistics, and has been done for many languages; two illustrative examples are English \citep{englanddialects,lap} and Japanese \citep{japanesedialects}. Dialect mapping is instrumentally important for the study of historical--comparative linguistics, since the present-day geography of isoglosses is a result of past \textit{language change} and \textit{language contact}. The distribution of synchronic features is data for theories of diachronic language change.

Computational approaches to dialect geography have worked on many parts of the issue, including the compilation of broad databases of linguistic features \citep{wals,diacl}, dialect identification and clustering on modern social media corpora \citep{abdul-mageed-etal-2018-tweet,jones2015toward}, and statistical modelling of dialect groups (e.g. \citealp{murawaki-2020-latent}).

South Asia is a \textit{linguistic area} \citep{masica,bashir}, a region of typological convergence due to historical contact between speakers of languages of different families. Families represented in South Asia are Indo-European, Dravidian, Austroasiatic, Sino-Tibetan, and some unclassified isolates (Nihali, Kusunda, and Burushaski).

Visualisation of data for linguistic typology has a long history, beginning with the first lexical isogloss maps created by aggregating data from dialect surveys and with more recent work specifically for visualising historical change, such as \citet{kalouli-etal-2019-parhistvis}. As linguists adopt computational methods that deal with vast amounts of data, it becomes a challenge for humans to interpret datasets. Modern approaches to visualisation like Visual Analytics (VA) try to address this issue \citep{keim2008visual,maceachren2017leveraging}.

The use of point-based mapping in linguistic data visualisation is well-known, in e.g. WALS \citep{wals}. This format has been used to map data in South Asian languages \citep{retroflexion,liljegren} as well as the languages of Iran \citep{ali,ali2}. We develop this paradigm further to map \underline{areal} language extents based on the location data in published linguistic fieldwork.

\section{Data model}

\begin{figure}[t]
\centering
\includegraphics[width=0.5\textwidth]{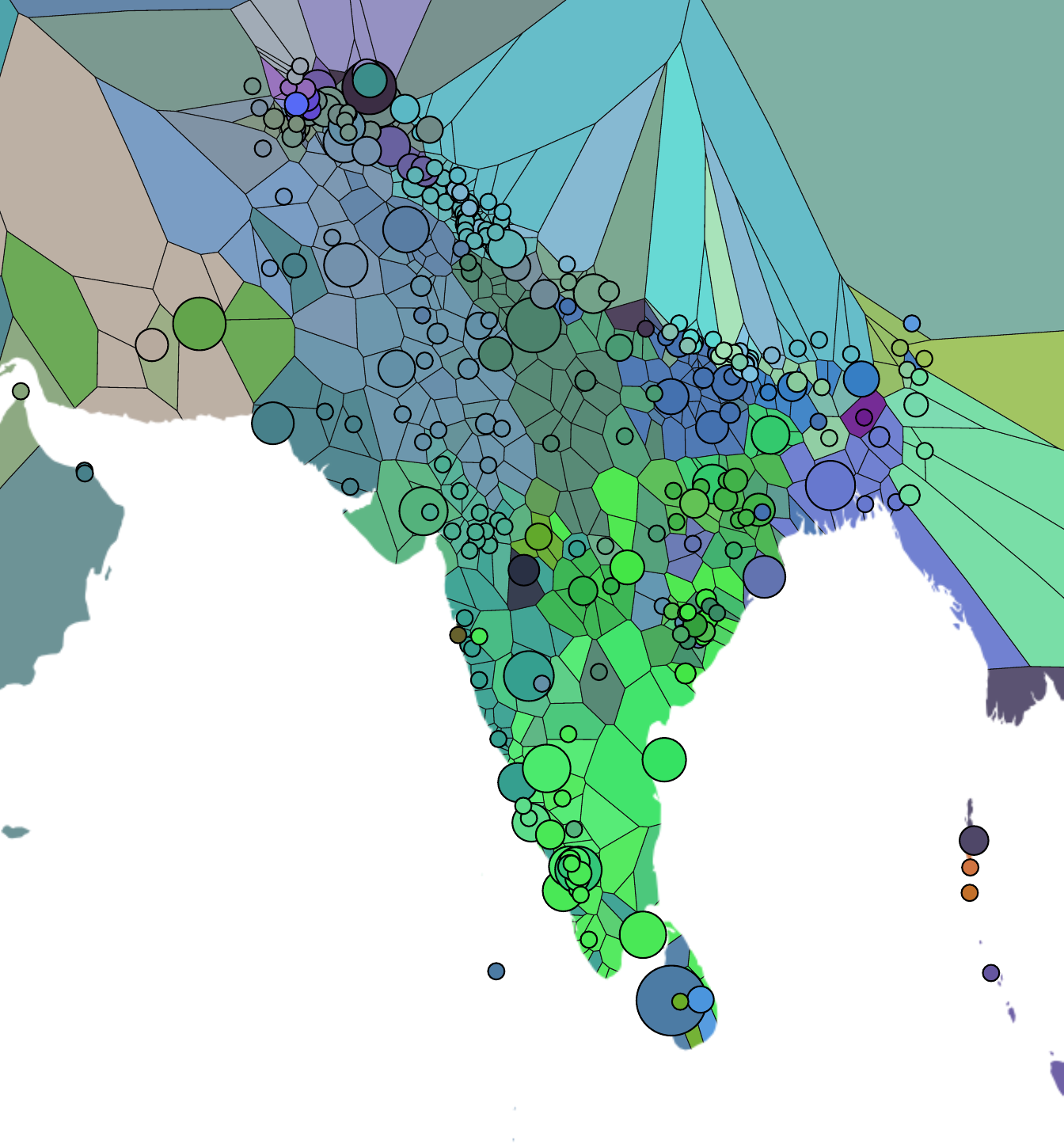}
\caption{\label{fig:map}The primary interface map for Bhāṣācitra generated with D3.js and Voronoi partitioning.}
\end{figure}

We built Bhāṣācitra to be an easy-to-use system for researchers with no computational background. We implemented the application in JavaScript on a statically-hosted webpage. There are three data files in JSON format, for reference metadata (in Bib\TeX{}-compatible format with additional fields for location and topic information; see \cref{sec:sources}), language metadata (traditional genetic classification and coordinates for reference locations), and the typological database (containing per-language per-location data).

The primary interface is an interactive map displaying geographical points corresponding to locations from which language data has been collected. The map is generated and manipulated using the D3.js library which has a complete pipeline for web cartography \citep{d3}. Dialect zones are partitioned using the Voronoi algorithm; for a point $P_k$ in the set of points $P$, its Voronoi region $R_k$ is defined as all points closer to $P_k$ than to any other point.
\begin{equation}
    R_k = \left\{x \in X \mathrel{\bigg|} \argmin_i{(\mathrm{dist}(P_i,\, x))} = k \right\}
\end{equation}

In the primary interface (see \cref{fig:map}), zones are colour-coded by consensus genetic classification of the languages covering the zone, with circles (with size proportional to the number of sources) centered at the weighted average of the coordinates of descriptions of the languages. In the case where multiple languages share a zone, the RGB components of the colouring are averaged.

\subsection{Interface}

\begin{figure}[t]
\centering
\includegraphics[width=0.5\textwidth]{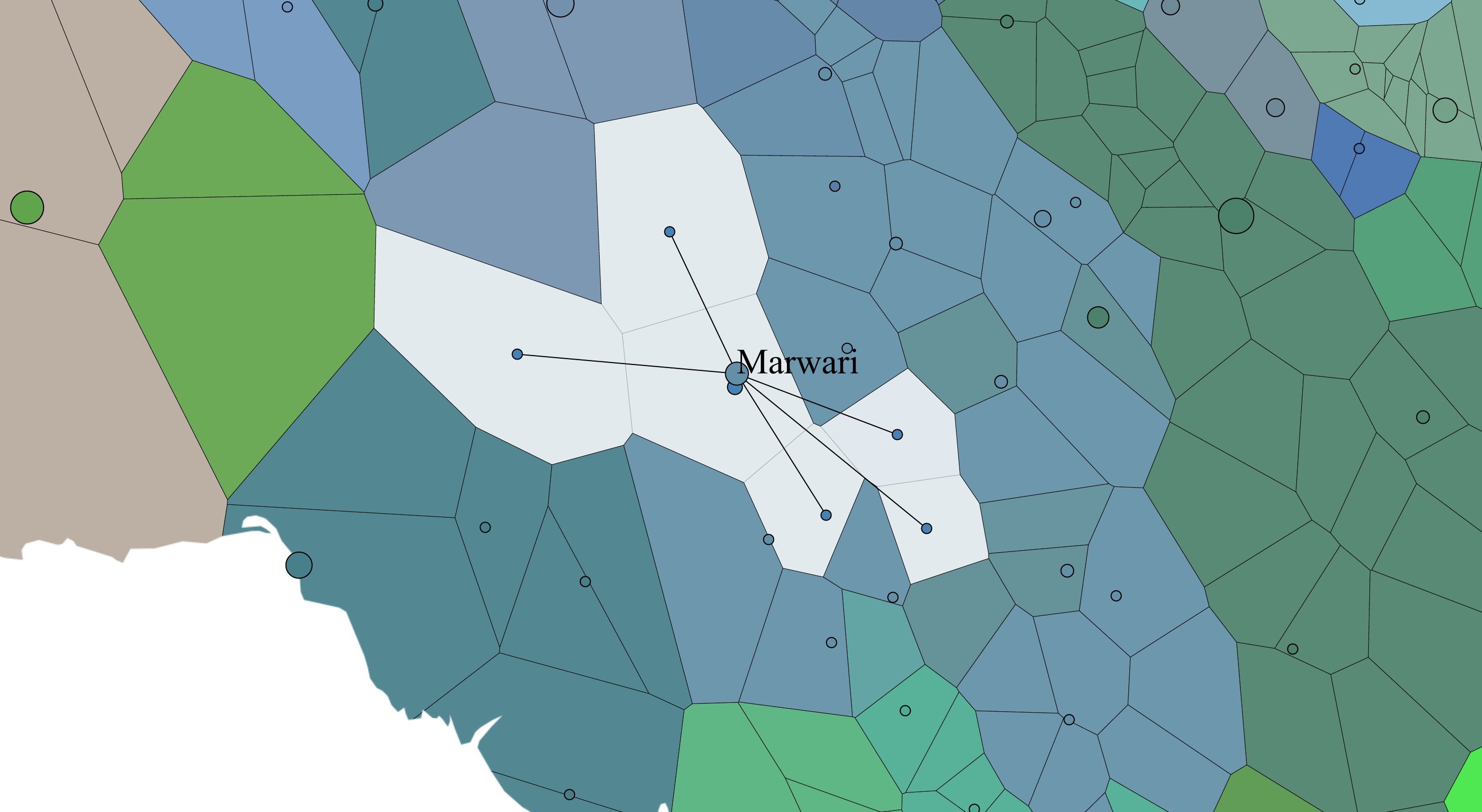}
\caption{\label{fig:marwari}Hovering on the circle for Marwari (a language of Rajasthan, India) highlights the regions from which linguistic sources for it draw data.}
\end{figure}

The primary interface map is fully interactive (draggable and zoomable). Hovering over a language circle shows all the geographical points and Voronoi polygons associated with the sources compiled for that language (see \cref{fig:marwari}). Like the language circles, each geographical point's size is weighted by the number of sources corresponding to it. Clicking on a language circle brings up the scrollable bibliography for that language, with each entry in human-readable format with the corresponding location and topic annotations appended.

\subsection{Limitations}
In South Asia (as elsewhere), geography is hardly the only variable encoding language use. As noted by \citet{deo2018dialects} and shown in sociolinguistic studies \citep{gumperz} factors such as caste, social status, political affiliation, and religion play a large role in language use and adoption. Migrant speaker communities have also developed distinct dialects even in regions where they are a minority language group (e.g.~Marathi speakers in Thanjavur and Burushaski speakers in Srinagar).

To deal with geographical overlap (different language sources for the same location), we allowed the areal zones of multiple languages to encompass the same location. A complete solution to the limitations of the geographical model would require collection of demographic data indexed to language use, which has not yet been collected on a large scale in South Asia.

\section{Compiling the database}

\begin{table}
    \centering
    {\small
    \begin{tabular}{lr}
        \toprule
        \textbf{Topic} & \textbf{Count} \\
        \midrule
        overview (descriptive grammars) & 494 \\
        syntax & 141 \\
        phonetics/phonology & 125 \\
        historical & 111 \\
        morphology & 100 \\
        sociolinguistics & 91 \\
        lexicography & 83 \\
        corpora & 51 \\
        dialectology & 48 \\
        comparative & 44 \\
        \midrule
        \textit{Total} & 1104 \\
        \bottomrule
    \end{tabular}}
    \caption{Count of sources labelled under the top 10 topics. A single source can be labelled with multiple topics.
    }
    \label{tab:topics}
\end{table}

There are some existing bibliographies of language references for South Asia. In compiling data for Bhāṣācitra, we prioritised the incorporation of sources that provided the greatest coverage of language information, such as grammars and grammatical sketches, analysed corpora, and sociolinguistic surveys.

We began with data from Glottolog for broad coverage \citep{glottolog}; South Asia-specific sources we drew from are \citet{seldom,np,kp}. We then searched for literature not included in existing bibliographies. Many new sources were obtained from Shodhganga,\footnote{\url{https://shodhganga.inflibnet.ac.in/}} a platform for open-access digitised theses completed at Indian universities. These theses were difficult to access before the past decade, so from this resource we were able to incorporate many new references.

We annotated information on topic coverage for every source (see \cref{tab:topics}) and location data (see \cref{sec:location}) when possible. We also preferred to link to open-access versions of sources. In total, we compiled \textbf{1104 sources} describing \textbf{311 lects} with data collected from \textbf{763 locations}. This number is continually increasing as we actively improve our coverage of the linguistic literature and new work is published.

\subsection{Locations}\label{sec:location}

\begin{figure}[t]
\centering
\includegraphics[width=0.5\textwidth]{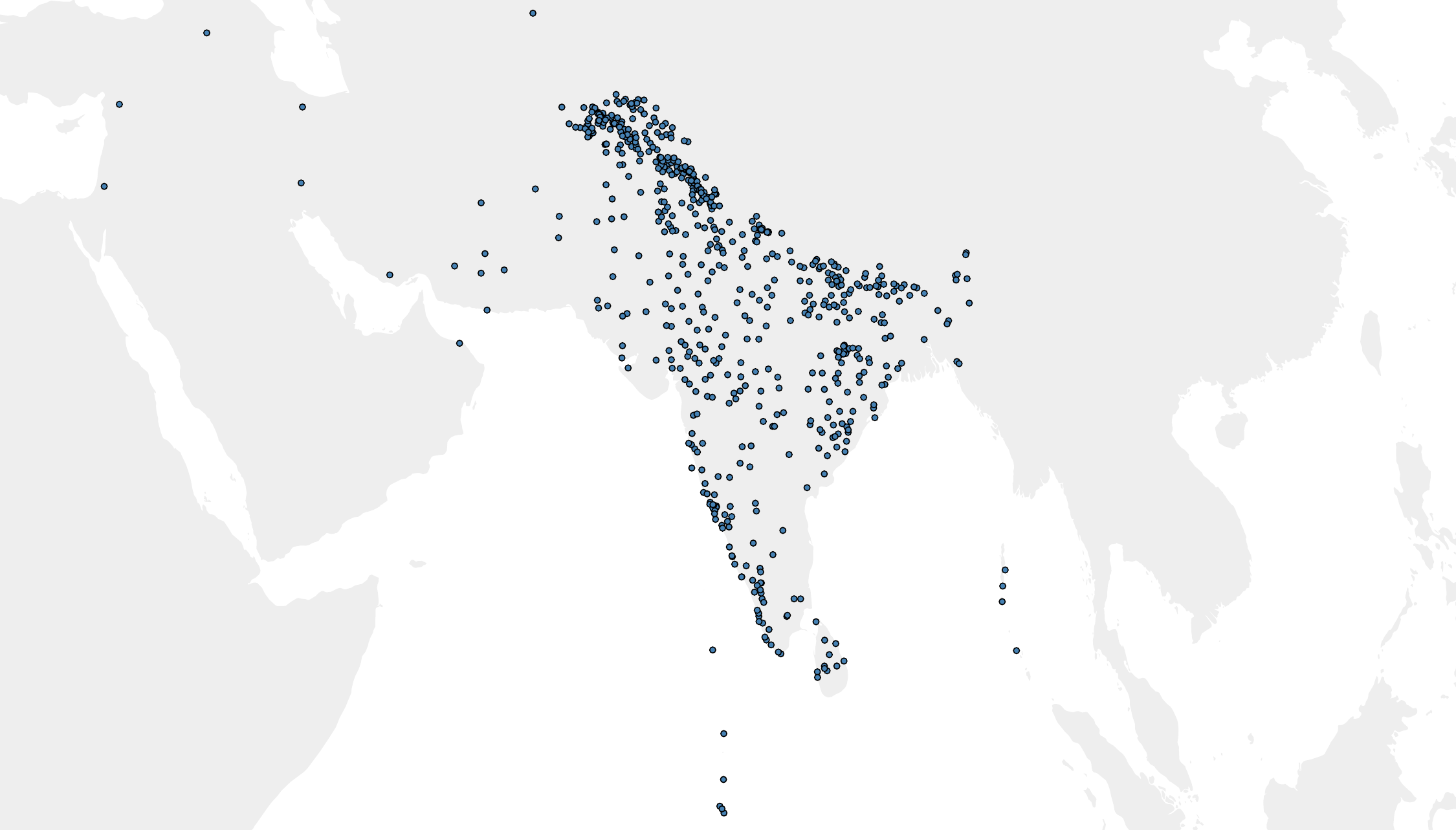}
\caption{\label{fig:map}Map of all locations extracted from the sources in the Bhāṣācitra database.}
\end{figure}

The primary new contribution of the Bhāṣācitra database is location data manually collected from the included references, shown in \cref{fig:map}. The geocoding of the locations was done through the Google Maps API and manually verified.

While databases such as Glottolog and WALS do include location data for languages, their representation reduces the language's geographical distribution to a single point. We instead represent multiple points per language based on data from the sources we catalogued.


For example, in Glottolog, Hindi is placed at a single point in central India, whereas in Bhāṣācitra there are 21 locations associated with Hindi--Urdu, with most sources describing the standard dialect in Delhi, but also work dealing with varieties in Varanasi, Lahore, and the rural regions surrounding Delhi. Areal mapping of linguistic references allows for better assessment of the coverage of dialects in our sources, and for explicit coverage of dialect variation when mapping features.

\begin{figure*}[h]
\centering
\begin{subfigure}[b]{0.43\textwidth}
\centering
\includegraphics[width=0.9\textwidth]{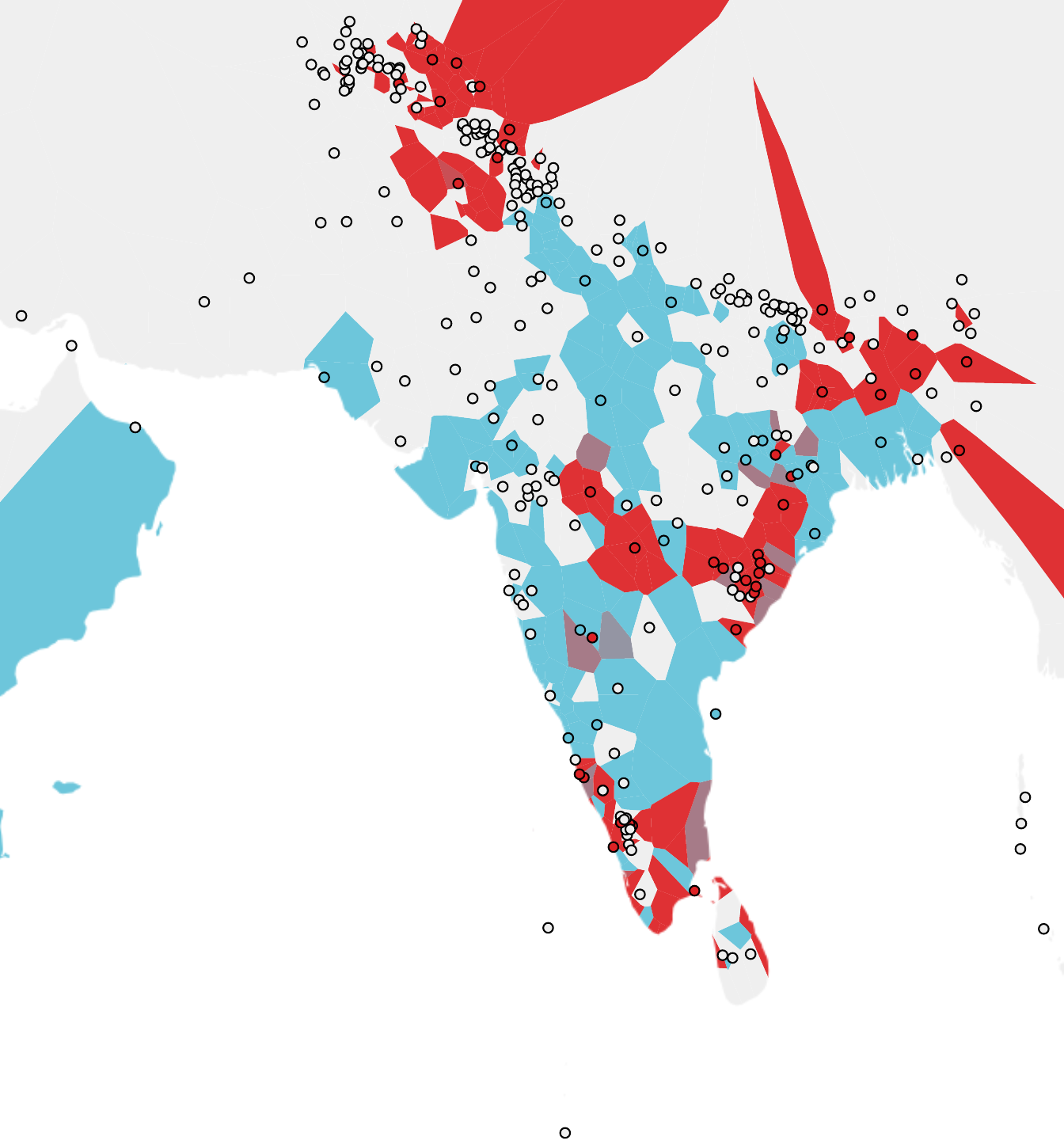}
\caption{\label{fig:dh}Distribution of the breathy-voiced retroflex stop (/\textipa{\:d\super H}/) in South Asian languages.}
\end{subfigure}
\begin{subfigure}[b]{0.43\textwidth}
\centering
\includegraphics[width=0.9\textwidth]{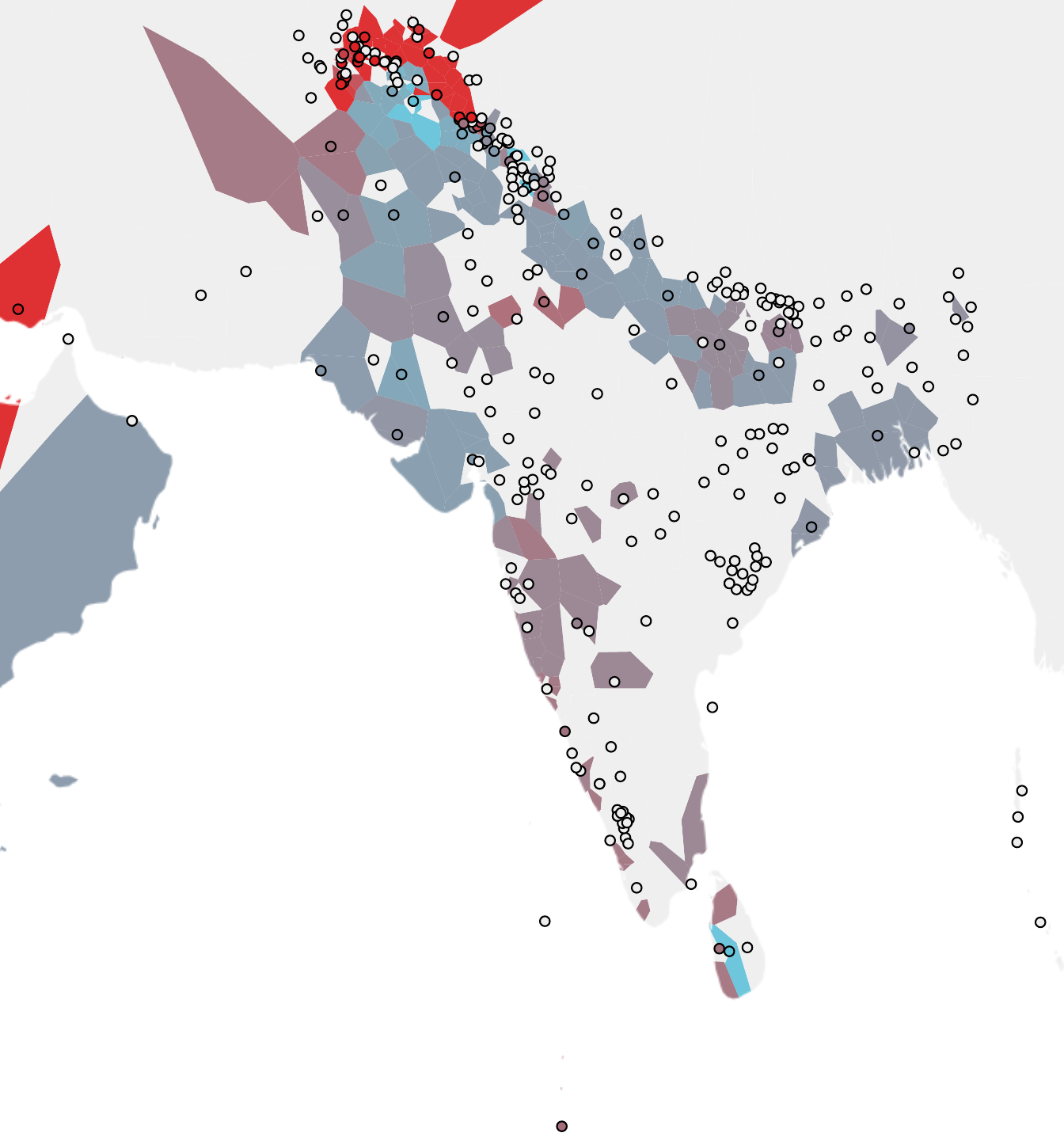}
\caption{\label{fig:ks}Percent of sound changes of Sanskrit \textipa{/k\:s/} that result in \textipa{/k(:)\super{h}/} in various Indo-Aryan languages.}
\end{subfigure}
\definecolor{green}{HTML}{63c2d8}
\definecolor{red}{HTML}{dd2225}
\caption{\label{fig:figs}Example datasets mapped in Bhāṣācitra. Scale: \fcolorbox{black}{green}{\rule{0pt}{6pt}\rule{6pt}{0pt}} Yes/100\%, \fcolorbox{black}{red}{\rule{0pt}{6pt}\rule{6pt}{0pt}} No/0\%}
\end{figure*}

\section{Mapping datasets}

To illustrate the value of areal visualisation of language features, we mapped two datasets: the phoneme inventories of a large number of Indian languages from \citet{ramaswami1999}, and the outcomes of selected sound changes from Sanskrit to the modern Indo-Aryan languages based on the Jambu database \citep{jambu} parsed from \citet{CDIAL}.

Note that we only visually analyse the map in these examples; these observations would need to be corroborated with statistic analysis and modelling to result in any verifiable claims.

\subsection{Phoneme inventories}

From the data in \citet{ramaswami1999} collected in the PHOIBLE database \citep{phoible} we were able to map the phoneme inventories of 62 major South Asian languages. Several works have studied the phonetic typology of the South Asian linguistic area, e.g.~\citet{phonetictypology,retroflexion}, but have not used areal mapping visualisations. 

Some interesting phonological features for mapping are retroflexion (which is prevalent throughout the region, but weakly distinguished or not distinguished at all in the eastern periphery) and breathy-voiced stops (which are less common in much of the Dravidian and Munda families and in the northwestern languages). \Cref{fig:dh} shows the distribution of the breathy-voiced retroflex stop /\textipa{\:d\super H}/ (in IAST: \textit{\d{d}h}) using the Bhāṣācitra system.

While \citet{retroflexion} did use mapping, the feature-separating lines were calculated based on point coordinates for each language, not areal zones. Bhāṣācitra produces more accurate visualisations; it is immediately clear that the northwest Indo-Aryan and Nuristani, Dravidian, and Munda languages lack the phoneme, and this information can be used to inform locations for future fieldwork at the isogloss boundaries to refine our data.

\subsection{Indo-Aryan sound changes}

As another demonstration, we use an under-development etymological database of Indo-Aryan languages \citep{jambu} that builds on \citet{CDIAL} to map the outcomes of some key Indo-Aryan sound changes.\footnote{The compilation of the Jambu database is not in the scope of this work, but, briefly, it has been compiled by parsing data from the digitised version of \citet{CDIAL} and augmenting it with several more recent diachronic dictionaries for Indo-Aryan languages.}

The Indo-Aryan (IA) languages show complex overlapping phonological isoglosses as a symptom of intense cross-dialectal contact over a long period of time, whose complexity makes it difficult to make sense of the family's linguistic history. For example, the Sanskrit cluster \textipa{/k\:s/} generally develops to \textipa{/k\super{h}/} in the core region of modern Indo-Aryan and \textipa{/\t{tS}\super{h}/} in the periphery, but some doublets are evidence of dialect contact, e.g. Sanskrit \textipa{/k\:sa:r@/} > Hindi \textipa{/\t{tS}\super{h}a:r/} `ashes' as well as \textipa{/k\super{h}a:r/} `alkali' \citep{masica}. The variability of these sound changes has recently been used to statistically model dialect components in IA languages \citep{cathcart-2019-gaussian,cathcart-2019-toward,cathcart-probabilistic,cathcart-rama-2020-disentangling}.

Thus, a visualisation of the probability of certain IA sound changes based on a lexical database would be useful for finding isoglosses and the geographical extent of historical dialect contact. We aligned the cognate forms given in \citet{jambu} using the LingPy library's multiple alignment function \citep{lingpy}. Based on the alignments, the likelihood of \textipa{/k\:s/} > \textipa{/k(:)\super{h}/} is mapped in \cref{fig:ks}. A rough core--periphery distinction indeed emerges, with languages in the northwest, south, and east having fewer outcomes of \textipa{/k(:)\super{h}/}. It is also apparent that the language coverage in \citet{CDIAL} is limited, with a great deal of core IA languages lacking data.

\section{Future work}

We intend to maximise coverage of South Asian languages in Bhāṣācitra. In the interest of achieving this goal we welcome contributions to our open-source database on GitHub: \url{https://github.com/aryamanarora/bhasacitra}. Ultimately, this sort of database would be useful for all languages of the world, but we lack the domain knowledge for non-South Asian languages, so we welcome any collaborators who feel this system would be beneficial.

As for directions for technical work, Bhāṣācitra would benefit from a SQL database for faster querying and precomputation of some data (e.g.~language circle sizes and coordinates) to improve performance in the browser. In the interface, we will explore continuous alternatives to discretised Voronoi polygons, which force rigid transitions between lects\footnote{We thank both reviewers for pointing out this limitation.} and do not show where location coverage is sparse. This will also help us with the issue of large polygons at the edges of our research area. Also, a basemap with administrative boundaries and other contextual geographical information would be useful. All of these will require substantial changes to the code beyond the capabilities of visualisation with pure D3.js.

Bhāṣācitra is one step of our larger goal of improving the study of South Asian languages with computational methods. Our future work on historical/comparative linguistics \citep{jambu} and corpus linguistics for under-studied languages of the region will benefit from Bhāṣācitra's visualisation capabilities.

\section{Conclusion}

We developed and presented Bhāṣācitra, a database of linguistic resources for South Asia and a language visualisation system based on location data from those resources. We analysed the coverage of our database and used the areal mapping system to visualise phoneme inventories and Indo-Aryan sound change outcomes. We hope that researchers find the tool useful especially as we move forward with studying the typology of South Asian languages.

\section*{Acknowledgments}

We thank Kaushalya Perera for providing her personal linguistic bibliography for Sinhala, Erik Anonby for showing us the \textit{Atlas of the Languages of Iran} (ALI) project, and Henrik Liljegren for pointing us to his work on Hindu Kush typology.

We also thank Nathan Schneider for his helpful comments on the paper and Ananya Chakravarti for the useful discussion when devising this project.

\bibliography{acl2020}

\begin{thebibliography}{37}
\expandafter\ifx\csname natexlab\endcsname\relax\def\natexlab#1{#1}\fi

\bibitem[{Abdul-Mageed et~al.(2018)Abdul-Mageed, Alhuzali, and
  Elaraby}]{abdul-mageed-etal-2018-tweet}
Muhammad Abdul-Mageed, Hassan Alhuzali, and Mohamed Elaraby. 2018.
\newblock \href {https://www.aclweb.org/anthology/L18-1577} {You tweet what you
  speak: A city-level dataset of {A}rabic dialects}.
\newblock In \emph{Proceedings of the Eleventh International Conference on
  Language Resources and Evaluation ({LREC} 2018)}, Miyazaki, Japan. European
  Language Resources Association (ELRA).

\bibitem[{Anonby et~al.(2019)Anonby, Taheri-Ardali, and Hayes}]{ali}
Erik Anonby, Mortaza Taheri-Ardali, and Amos Hayes. 2019.
\newblock \href {https://doi.org/10.1080/00210862.2019.1573135} {The {A}tlas of
  the {L}anguages of {I}ran ({ALI}): A research overview}.
\newblock \emph{Iranian studies}, 52(1--2):199--230.

\bibitem[{Anonby et~al.(2018)Anonby, Taheri-Ardali, Taylor, and Hayes}]{ali2}
Erik Anonby, Mortaza Taheri-Ardali, Fraser Taylor, and Amos Hayes. 2018.
\newblock \href {https://ir.library.carleton.ca/pub/17913/} {Atlas of the
  {L}anguages of {I}ran}.

\bibitem[{Arora and Farris(2021)}]{jambu}
Aryaman Arora and Adam Farris. 2021.
\newblock \href {http://jambu-clld.herokuapp.com/} {\emph{Jambu}}.
\newblock Georgetown University, Washington.

\bibitem[{Arsenault(2017)}]{retroflexion}
Paul Arsenault. 2017.
\newblock \href {https://doi.org/doi:10.1515/jsall-2017-0001} {Retroflexion in
  {S}outh {A}sia: Typological, genetic, and areal patterns}.
\newblock \emph{Journal of South Asian Languages and Linguistics}, 4(1):1--53.

\bibitem[{Asher(2008)}]{asher2008language}
Ronald~E Asher. 2008.
\newblock Language in historical context.
\newblock \emph{Language in South Asia}, pages 31--48.

\bibitem[{Baart and Baart-Bremer(2001)}]{np}
Joan L.~G. Baart and Esther~L. Baart-Bremer. 2001.
\newblock \href {https://www.sil.org/resources/archives/38577}
  {\emph{Bibliography of languages of northern {P}akistan}}.
\newblock NIPS--SIL Working Paper Series. National Institute of Pakistan
  Studies, Quaid-i-Azam University and Summer Institute of Linguistics.

\bibitem[{Bashir(2016)}]{bashir}
Elena Bashir. 2016.
\newblock Contact and convergence.
\newblock In Hans~Henrich Hock and Elena Bashir, editors, \emph{The Languages
  and Linguistics of {S}outh {A}sia: A Comprehensive Guide}. De Gruyter Mouton.

\bibitem[{Bostock et~al.(2011)Bostock, Ogievetsky, and Heer}]{d3}
Michael Bostock, Vadim Ogievetsky, and Jeffrey Heer. 2011.
\newblock \href {https://doi.org/10.1109/TVCG.2011.185} {D³ data-driven
  documents}.
\newblock \emph{IEEE Transactions on Visualization and Computer Graphics},
  17(12):2301--2309.

\bibitem[{Carling et~al.(2018)Carling, Larsson, Cathcart, Johansson, Holmer,
  Round, and Verhoeven}]{diacl}
Gerd Carling, Filip Larsson, Chundra~A. Cathcart, Niklas Johansson, Arthur
  Holmer, Erich Round, and Rob Verhoeven. 2018.
\newblock \href {https://doi.org/10.1371/journal.pone.0205313} {{D}iachronic
  {A}tlas of {C}omparative {L}inguistics ({DiACL})—a database for ancient
  language typology}.
\newblock \emph{PLOS ONE}, 13(10):1--20.

\bibitem[{Cathcart(2019{\natexlab{a}})}]{cathcart-2019-gaussian}
Chundra Cathcart. 2019{\natexlab{a}}.
\newblock \href {https://doi.org/10.18653/v1/W19-4732} {{G}aussian process
  models of sound change in {I}ndo-{A}ryan dialectology}.
\newblock In \emph{Proceedings of the 1st International Workshop on
  Computational Approaches to Historical Language Change}, pages 254--264,
  Florence, Italy. Association for Computational Linguistics.

\bibitem[{Cathcart(2019{\natexlab{b}})}]{cathcart-2019-toward}
Chundra Cathcart. 2019{\natexlab{b}}.
\newblock \href {https://doi.org/10.18653/v1/W19-1411} {Toward a deep
  dialectological representation of {I}ndo-{A}ryan}.
\newblock In \emph{Proceedings of the Sixth Workshop on {NLP} for Similar
  Languages, Varieties and Dialects}, pages 110--119, Ann Arbor, Michigan.
  Association for Computational Linguistics.

\bibitem[{Cathcart(2020)}]{cathcart-probabilistic}
Chundra Cathcart. 2020.
\newblock A probabilistic assessment of the {I}ndo-{A}ryan {I}nner–{O}uter
  {H}ypothesis.
\newblock \emph{Journal of Historical Linguistics}, 10(1):42--86.

\bibitem[{Cathcart and Rama(2020)}]{cathcart-rama-2020-disentangling}
Chundra Cathcart and Taraka Rama. 2020.
\newblock \href {https://doi.org/10.18653/v1/2020.conll-1.50} {Disentangling
  dialects: a neural approach to {I}ndo-{A}ryan historical phonology and
  subgrouping}.
\newblock In \emph{Proceedings of the 24th Conference on Computational Natural
  Language Learning}, pages 620--630, Online. Association for Computational
  Linguistics.

\bibitem[{Deo(2018)}]{deo2018dialects}
Ashwini Deo. 2018.
\newblock Dialects in the {I}ndo-{A}ryan landscape.
\newblock In Charles Boberg, John Nerbonne, and Dominic Watt, editors,
  \emph{The Handbook of Dialectology}, pages 535--546. Wiley Online Library.

\bibitem[{Dryer and Haspelmath(2013)}]{wals}
Matthew~S. Dryer and Martin Haspelmath, editors. 2013.
\newblock \href {https://wals.info/} {\emph{WALS Online}}.
\newblock Max Planck Institute for Evolutionary Anthropology, Leipzig.

\bibitem[{Eberhard et~al.(2021)Eberhard, Simons, and Fennig}]{ethnologue}
David~M. Eberhard, Gary~F. Simons, and Charles~D. Fennig, editors. 2021.
\newblock \href {http://www.ethnologue.com} {\emph{Ethnologue: Languages of the
  World}}, 24th edition.
\newblock SIL International.

\bibitem[{Gumperz(1958)}]{gumperz}
John~J. Gumperz. 1958.
\newblock Dialect differences and social stratification in a {N}orth {I}ndian
  village.
\newblock \emph{American Anthropologist}, 60(4):668--682.

\bibitem[{Hammarstr\"om et~al.(2020)Hammarstr\"om, Forkel, Haspelmath, and
  Bank}]{glottolog}
Harald Hammarstr\"om, Robert Forkel, Martin Haspelmath, and Sebastian Bank.
  2020.
\newblock \href {https://doi.org/10.5281/zenodo.4061162} {\emph{Glottolog
  4.3}}.
\newblock Max Planck Institute for the Science of Human History.

\bibitem[{Jones(2015)}]{jones2015toward}
Taylor Jones. 2015.
\newblock Toward a description of {A}frican {A}merican {V}ernacular {E}nglish
  dialect regions using ``{B}lack {T}witter''.
\newblock \emph{American Speech}, 90(4):403--440.

\bibitem[{Kalouli et~al.(2019)Kalouli, Kehlbeck, Sevastjanova, Kaiser, Kaiser,
  and Butt}]{kalouli-etal-2019-parhistvis}
Aikaterini-Lida Kalouli, Rebecca Kehlbeck, Rita Sevastjanova, Katharina Kaiser,
  Georg~A. Kaiser, and Miriam Butt. 2019.
\newblock \href {https://doi.org/10.18653/v1/W19-4714} {{P}ar{H}ist{V}is:
  Visualization of parallel multilingual historical data}.
\newblock In \emph{Proceedings of the 1st International Workshop on
  Computational Approaches to Historical Language Change}, pages 109--114,
  Florence, Italy. Association for Computational Linguistics.

\bibitem[{Keim et~al.(2008)Keim, Andrienko, Fekete, G{\"o}rg, Kohlhammer, and
  Melan{\c{c}}on}]{keim2008visual}
Daniel Keim, Gennady Andrienko, Jean-Daniel Fekete, Carsten G{\"o}rg, J{\"o}rn
  Kohlhammer, and Guy Melan{\c{c}}on. 2008.
\newblock Visual analytics: Definition, process, and challenges.
\newblock In \emph{Information visualization}, pages 154--175. Springer.

\bibitem[{Kretzschmar(2001)}]{lap}
William~A. Kretzschmar. 2001.
\newblock Linguistic databases of the {A}merican {L}inguistic {A}tlas {P}roject
  ({ALAP}).

\bibitem[{Kumagai(2016)}]{japanesedialects}
Yasuo Kumagai. 2016.
\newblock Developing the {L}inguistic {A}tlas of {J}apan {D}atabase and
  advancing analysis of geographical distributions of dialects.
\newblock In Marie-Hélène Côté, Remco Knooihuizen, and John Nerbonne,
  editors, \emph{The future of dialects: Selected papers from Methods in
  Dialectology XV}. Language Science Press.

\bibitem[{Liljegren et~al.(2021)Liljegren, Forkel, Knobloch, and
  Lange}]{liljegren}
Henrk Liljegren, Robert Forkel, Nina Knobloch, and Noa Lange. 2021.
\newblock \href {https://hindukush.clld.org/} {Hindu {K}ush areal typology
  (version v1.0)}.

\bibitem[{List et~al.(2019)List, Greenhill, Tresoldi, and Forkel}]{lingpy}
Johann-Mattis List, Simon~J Greenhill, Tiago Tresoldi, and Robert Forkel. 2019.
\newblock \href {https://doi.org/10.5281/zenodo.3554103} {{LingPy. A Python
  library for quantitative tasks in historical linguistics}}.

\bibitem[{MacEachren(2017)}]{maceachren2017leveraging}
Alan~M MacEachren. 2017.
\newblock Leveraging big (geo) data with (geo) visual analytics: Place as the
  next frontier.
\newblock In \emph{Spatial data handling in big data era}, pages 139--155.
  Springer.

\bibitem[{Masica(1993)}]{masica}
Colin~P. Masica. 1993.
\newblock \emph{The Indo-Aryan languages}.
\newblock Cambridge Lamguage Surveys. Cambridge University Press, Cambridge.

\bibitem[{Moran and McCloy(2019)}]{phoible}
Steven Moran and Daniel McCloy, editors. 2019.
\newblock \href {https://phoible.org/} {\emph{PHOIBLE 2.0}}.
\newblock Max Planck Institute for the Science of Human History, Jena.

\bibitem[{Murawaki(2020)}]{murawaki-2020-latent}
Yugo Murawaki. 2020.
\newblock \href {https://doi.org/10.18653/v1/2020.emnlp-main.69} {Latent
  geographical factors for analyzing the evolution of dialects in contact}.
\newblock In \emph{Proceedings of the 2020 Conference on Empirical Methods in
  Natural Language Processing (EMNLP)}, pages 959--976, Online. Association for
  Computational Linguistics.

\bibitem[{Orton et~al.(1998)Orton, Sanderson, and Widdowson}]{englanddialects}
Harold Orton, Stewart Sanderson, and John Widdowson, editors. 1998.
\newblock \emph{The linguistic atlas of {E}ngland}.
\newblock Psychology Press.

\bibitem[{Perera(2021)}]{kp}
Kaushalya Perera. 2021.
\newblock Personal communication.

\bibitem[{Peterson(2018)}]{seldom}
John Peterson. 2018.
\newblock \href
  {https://www.isfas.uni-kiel.de/de/linguistik/forschung/projekte/southasiabibliography}
  {Bibliography for seldom studied and endangered {S}outh {A}sian languages}.

\bibitem[{Ramanujan and Masica(2016)}]{phonetictypology}
A.~K. Ramanujan and Colin Masica. 2016.
\newblock \href {https://doi.org/doi:10.1515/9783110819502-029} {\emph{Toward a
  Phonological Typology of the {I}ndian Linguistic Area}}, pages 543--577. De
  Gruyter Mouton.

\bibitem[{Ramaswami(1999)}]{ramaswami1999}
N.~Ramaswami. 1999.
\newblock \href {https://phoible.org/contributors/RA} {\emph{Common Linguistic
  Features in Indian Languages: Phonetics}}.
\newblock Central Institute of Indian Languages.

\bibitem[{Shackle(1980)}]{hindko}
Christopher Shackle. 1980.
\newblock \href {https://www.jstor.org/stable/615737} {{H}indko in {K}ohat and
  {P}eshawar}.
\newblock \emph{Bulletin of the School of Oriental and African Studies,
  University of London}, 43(3):482--510.

\bibitem[{Turner(1962–1966)}]{CDIAL}
Ralph~Lilley Turner. 1962–1966.
\newblock \href {https://dsal.uchicago.edu/dictionaries/soas/} {\emph{A
  comparative dictionary of the Indo-Aryan languages}}.
\newblock Oxford University Press.

\end{thebibliography}
\bibliographystyle{acl_natbib}

\appendix

\section{Source format}\label{sec:sources}

Below is reference metadata for \citet{hindko} in JSON format; note the location annotations and the topic data.

\small
\begin{verbatim}
{
    "type": "article",
    "title": "Hindko in Kohat and Peshawar",
    "author": ["Christopher Shackle"],
    "journal": "Bulletin of the School...",
    "year": 1980,
    "volume": 43,
    "number": 3,
    "pages": "482--510",
    "url": "https://www.jstor.org/stable/615737",
    "languages": {
        "Hindko": ["Kohat", "Peshawar"]
    },
    "topics": ["overview"]
}
\end{verbatim}




\end{document}